\documentclass{article}

\PassOptionsToPackage{numbers}{natbib}

\usepackage[main, final]{neurips_2026}
\usepackage[utf8]{inputenc} 
\usepackage[T1]{fontenc}    
\usepackage{hyperref}       
\usepackage{url}            
\usepackage{booktabs}       
\usepackage{amsmath}        
\usepackage{amsfonts}       
\usepackage{nicefrac}       
\usepackage{microtype}      
\usepackage{xcolor}         
\usepackage{graphicx}       
\title{Beyond Waypoints: Dual-Heatmap Grounding for Cross-Embodiment Semantic Navigation}

\author{%
  Kaijie Yun\\
  Harbin Institute of Technology\\
  \texttt{120L030418@stu.hit.edu.cn} \\
  \And
  Yue Chen\thanks{Corresponding author.} \\
  JD AI Research \\
  \texttt{chenyue21@jd.com} \\
}

\begin{document}

\maketitle

\begin{abstract}
Grounding open-ended semantic instructions into physically executable local goals is a fundamental challenge in human-robot interaction. While existing navigation frameworks often regress deterministic waypoints, this rigid formulation collapses spatial uncertainty and frequently targets non-traversable object centers, leading to severe execution failures. In this work, we focus on the practical setting of in-FOV semantic navigation, where a robot receives concise, interleaved multimodal (text and image) prompts. To bridge the gap between abstract semantic intent and physical reachability, we propose a unified Vision-Language framework that abandons single-point regression in favor of a Dual-Heatmap representation. Our framework predicts a navigation affordance heatmap that captures continuous reachable regions, coupled with a facing heatmap for orientation constraints. These dense outputs inherently function as a differentiable semantic potential field, integrating seamlessly with downstream local planners. To support this paradigm, we build a fully automated, foundation-model-assisted synthetic data pipeline and establish a comprehensive simulation benchmark. Extensive experiments demonstrate that our framework achieves state-of-the-art performance among comparable 8B baselines. Crucially, a feature-fusion study and simulation studies across diverse robot embodiments (Jetbot, H1, Aliengo) reveal that explicit heatmap prediction drastically improves the Affordance Rate (AR). By placing targets reliably in executable free space, our framework effectively mitigates the brittleness of point regression, offering a transferable path toward safe cross-embodiment semantic navigation.
\end{abstract}

\section{Introduction}

Environmental perception and semantic understanding are fundamental capabilities for autonomous robots operating in human-centric spaces. While traditional navigation systems excel at traversing metric maps using geometric coordinates, human-robot interaction fundamentally relies on abstract, semantic concepts. When a user commands a robot to ``go to the sofa'' or ``approach the user'', the robot's perception module must act as a crucial bridge, grounding high-level semantic intentions into physically actionable anchors.

Historically, this semantic grounding has been addressed through a modular pipeline relying on closed-set perception models, such as category-specific object detectors or semantic segmentation networks. While effective in constrained environments, these models suffer from limited generalization capability. Recently, the advent of Vision-Language Models (VLMs) and open-vocabulary foundation models has driven a paradigm shift. By enabling zero-shot recognition and natural language grounding, modern robotic systems can now interpret open-ended user instructions.

However, existing frameworks that leverage language for navigation, most notably Vision-and-Language Navigation (VLN), often misalign with realistic deployment scenarios. VLN tasks typically assume the user provides a detailed, turn-by-turn linguistic route instruction to reach a distant goal. In practical robotic applications, such granular ``roadbooks'' are rarely available. Instead, real-world tasks often consist of concise, intent-driven commands aimed at targets within the robot's current or immediate sensor Field of View (FOV). Furthermore, while VLN models attempt an end-to-end mapping from pixels to control actions, they often struggle with the Sim-to-Real gap and lack the safety guarantees required for physical deployment.

Focusing on the realistic task of FOV-target semantic navigation, a critical challenge remains: the gap between semantic identification and physical reachability. Simply utilizing a VLM to generate a 2D bounding box or a segmentation mask of the target object is insufficient for navigation. Extracting a geometric waypoint from the center of a ``sofa'' mask inevitably leads to collisions, as the semantic center of an object is rarely a navigable space. Moreover, complex user intents often couple positional goals with orientational constraints—such as ``look out the window'' or ``accompany this person represented by \texttt{\textless person\_image\textgreater} to watch TV''—which cannot be represented by a simple geometric coordinate.

To address these challenges and bridge the gap between abstract semantic intent and physical reachability, we define a highly flexible, intent-driven semantic navigation task and propose a unified Vision-Language framework to solve it. As illustrated in Figure~\ref{fig:overall}, our system goes beyond rigid predefined commands by supporting a rich spectrum of multimodal inputs. Specifically, the instruction prompts can seamlessly take the following forms:
\begin{figure}[t]
  \centering
  \includegraphics[width=0.95\linewidth]{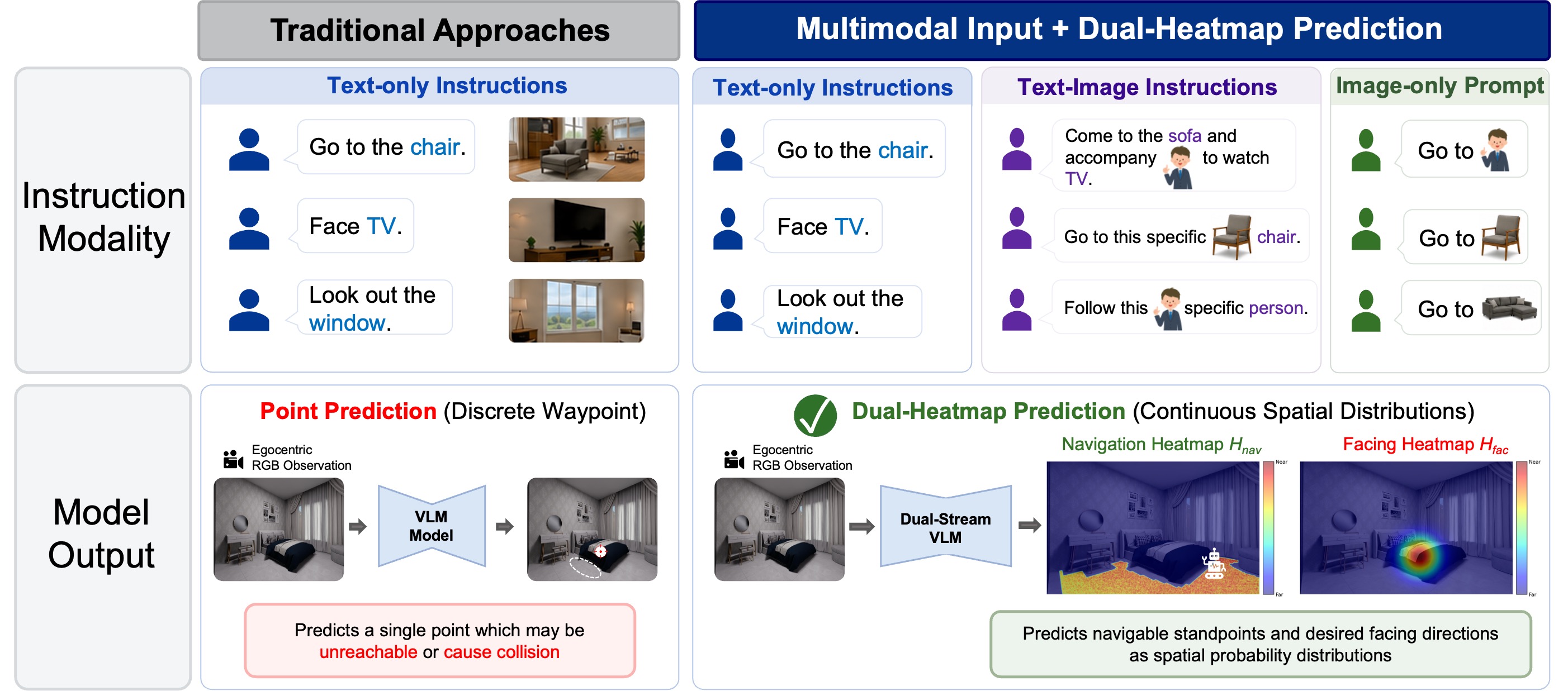}
  \caption{Overview of the proposed semantic navigation framework and supported multimodal instruction formats.}
  \label{fig:overall}
\end{figure}

\begin{itemize}

        \item {\bfseries Text-only instructions:} Specifying semantic targets or relative spatial relationships (e.g., ``go to the chair'', ``face the TV'', or ``look out the window'').
        \item {\bfseries Image-only prompts:} Providing a reference image of a target, allowing the robot to navigate directly to a specific user or a unique piece of furniture (e.g., navigating to user A given \texttt{\textless person\_image\_A\textgreater}, or to a specific chair given \texttt{\textless chair\_image\textgreater}).
        \item {\bfseries Interleaved text-image combinations:} Enabling complex, context-aware instructions that combine linguistic logic with visual constraints (e.g., ``come to the sofa and accompany \texttt{\textless person\_image\_A\textgreater} to watch TV'', or ``go to this specific \texttt{\textless chair\_image\textgreater}'').
\end{itemize}

Rather than translating these diverse and complex inputs into deterministic, discrete waypoints that may be physically unreachable due to dynamic obstacles or kinematic constraints, our model operates directly in the first-person 2D image space to predict a Dual-Heatmap representation. Specifically, given the current egocentric RGB observation and the aforementioned interleaved instruction, the VLM generates two probabilistic heatmaps: a navigation heatmap ($H_{\mathrm{nav}}$) and a facing heatmap ($H_{\mathrm{fac}}$).

This design draws inspiration from two distinct domains. First, akin to affordance maps in robotic manipulation—which predict ``where to grasp'' rather than just segmenting the object—our $H_{\mathrm{nav}}$ predicts the navigable ``standpoint affordance'' (e.g., the empty floor space adjacent to the sofa). Second, akin to costmaps in traditional robotic navigation, outputting a spatial probability distribution allows for an elegant decoupling of semantic decision-making from physical execution. The generated heatmaps act as an attractive potential field, seamlessly integrating with off-the-shelf local planners to dynamically compute the optimal, collision-free trajectory and the final yaw angle governed by $H_{\mathrm{fac}}$.
In summary, the main contributions of this work are as follows:
\begin{itemize}
        \item We introduce a novel open-vocabulary semantic navigation framework designed for realistic, in-FOV target anchoring, supporting highly flexible interleaved multimodal inputs (text and reference images).
        \item We propose an innovative Dual-Heatmap representation ($H_{\mathrm{nav}}$ and $H_{\mathrm{fac}}$) in the first-person perspective, elegantly bridging high-level VLM semantic reasoning with low-level costmap-based obstacle avoidance systems, ensuring both semantic accuracy and physical reachability.
        \item We present a comprehensive, automated data generation pipeline leveraging simulation engines and foundation model distillation to construct a large-scale dataset of paired multimodal instructions and affordance heatmaps.
        \item We demonstrate the effectiveness and robustness of our framework through extensive simulation-based validation across different robot embodiments, showing strong performance in complex human-centric environments.
\end{itemize}

\section{Related Work}

\subsection{Vision-Language Models in Embodied AI}
The emergence of Large Language Models (LLMs) and Vision-Language Models (VLMs) has fundamentally transformed embodied AI, shifting the paradigm from task-specific training to open-vocabulary, zero-shot reasoning. Foundation models, such as LLaVA\cite{liu2023llava}, Qwen-VL\cite{Qwen-VL}, and GPT-4V\cite{achiam2023gpt}, have demonstrated remarkable capabilities in visual understanding and common-sense spatial reasoning. In the robotics domain, pioneering works like PaLM-E\cite{driess2023palm} and RT-2\cite{zitkovich2023rt} have successfully integrated VLMs as high-level cognitive ``brains'' to directly map multimodal inputs into robotic actions or executable code policies. Furthermore, approaches like VoxPoser\cite{huang2023voxposer} extract 3D value maps from LLMs to guide manipulation without extensive retraining. While these models excel in high-level planning and tabletop manipulation, their direct application in mobile robot local navigation remains underexplored. Existing VLM-based mobile systems often rely on coarse spatial grounding or output deterministic coordinate commands. Our work bridges this gap by leveraging the dense prediction capabilities of VLMs to generate pixel-level probabilistic representations, explicitly designed to interface with lower-level dynamic local planners.

\subsection{Vision-and-Language Navigation (VLN)}
The field of VLN has evolved significantly, transitioning from navigating discrete topological graphs to operating in highly realistic continuous environments (VLN-CE), a shift pioneered by Krantz et al. \cite{krantz2020beyond, krantz2021waypoint}. More recently, Wang et al. \cite{wang2025rethinking} highlighted the importance of addressing the ``embodied gap,'' emphasizing the physical disparities between simulation and real-world deployment. Current methodologies for solving VLN and ObjectNav tasks typically fall into two categories: map-based reasoning and map-less reactive exploration.

Map-based methods rely on pre-built or incrementally constructed global maps. Huang et al. \cite{huang2023visual} utilized global BEV (Bird's Eye View) maps to store visual-language embeddings, while others \cite{chen2022weakly} proposed multi-granularity map learning to bridge instructions with structured layouts. To enhance scalability, hierarchical representations and 3D scene graphs have become dominant trends \cite{rana2023sayplan, werby2024hierarchical, chen2025astra, zhou2025fsr}, allowing robots to ground instructions into global metric structures. In contrast, map-less reactive approaches \cite{yu2023l3mvn, gadre2023cows, ramakrishnan2022poni, wei2025ground} leverage foundation models to identify target objects and frontiers in unknown environments for real-time exploration.

However, a common bottleneck in both paradigms is their reliance on predicting deterministic, discrete waypoints for execution. Predicting a rigid 3D coordinate or a single 2D pixel as a local goal forces a ``hard constraint'' on the system; if that specific point is obstructed by a dynamic obstacle or violates kinematic limits, the navigation stack often deadlocks. Unlike these general-purpose frameworks that focus on long-range exploration and waypoint prediction, our work addresses in-FOV target anchoring. We transform multimodal intent into a dense, first-person navigable potential field, inheriting semantic reasoning while avoiding the brittleness of deterministic waypoints.

\subsection{Spatial Grounding and Affordance Representation}
Spatial grounding is the process of localizing semantic entities in a physical or visual space based on linguistic descriptions. Traditional approaches relied on object detection or semantic segmentation to produce 2D bounding boxes or precise object masks. However, as noted in our introduction, these boundary-focused representations often fail to capture physical ``reachability'' (e.g., the center of a sofa mask is not a navigable coordinate). 

To address this, the concept of affordance—originally defined by Gibson \cite{gibson1978ecological} as the action possibilities latent in an environment—has been revitalized. Modern frameworks have shifted the inquiry from ``what is this object'' to ``where can I interact with it.'' While semantic memory structures like VLMaps \cite{huang2023visual} and CLIP-Fields \cite{shafiullah2022clip} provide rich semantic density, they often lack the fine-grained ``standpoint'' logic required for safe navigation. 

Instead of extracting deterministic coordinates from bounded objects, recent advancements demonstrate that predicting probabilistic heatmaps offers a more robust grounding mechanism. Shao et al. \cite{shao2025more} utilized adaptive affordance heatmaps to capture spatial uncertainty in robotic tasks, while Chen et al. \cite{chen2025affordances} showed that affordance-oriented planning with foundation models significantly enhances success rates in continuous environments. Heatmaps act as a ``soft constraint,'' functioning similarly to a semantic costmap. Our work extends this philosophy by introducing a Dual-Heatmap structure that simultaneously models the affordance of the navigation standpoint ($H_{\mathrm{nav}}$) and the directional constraint of the final orientation ($H_{\mathrm{fac}}$), seamlessly translating complex user intents into probability distributions that are natively compatible with traditional local planners.

\section{Method}

\subsection{Task Definition}
We define semantic navigation as a dense prediction problem in image space. Given the current RGB observation from a forward-facing camera,
\begin{equation}
I \in \mathbb{R}^{H \times W \times 3},
\end{equation}
and a multimodal user instruction $\mathcal{C}$, the model predicts two heatmaps with the same spatial resolution:
\begin{equation}
\{H_{\mathrm{nav}}, H_{\mathrm{fac}}\} \in \mathbb{R}^{H \times W}.
\end{equation}
Here, $H_{\mathrm{nav}}$ encodes the spatial likelihood of physically reachable navigation points, and $H_{\mathrm{fac}}$ encodes orientation intent toward a facing target.

The instruction $\mathcal{C}$ supports arbitrary interleaving of text and image references (e.g., ``Find and go to the sofa, then face the TV''). Both the navigation target and the facing target can be specified either by text labels or by reference image patches. If a target is absent from the current field of view, the corresponding heatmap is defined as an all-zero map (negative sample).

We choose dense heatmap prediction instead of directly regressing a single waypoint because semantic navigation is inherently multi-modal. Many instructions do not correspond to one isolated pixel-level destination, but to a set of semantically equivalent and physically acceptable regions. For example, the command to move toward a chair-bearing area or to accompany a person near a sofa may admit multiple valid standing locations. A point predictor must collapse this distribution into one coordinate, which introduces regression ambiguity and can cause unstable supervision when several solutions are equally correct. In contrast, the proposed heatmap navigation formulation preserves this spatial uncertainty as a dense affordance field.

This dense output is fundamentally more compatible with downstream feasibility filtering. By predicting a full spatial distribution, the model exposes a set of high-confidence candidate regions, after which the planner can combine the semantic signal with collision costs and embodiment constraints to select a feasible execution target. In this sense, the heatmap can be interpreted as a differentiable semantic potential field $U_{\mathrm{sem}}(x)$, and the local optimization objective can be formulated as:
\begin{equation}
J(x)=\alpha H_{\mathrm{nav}}(x)+\beta H_{\mathrm{fac}}(x)-\mathrm{Cost}_{\mathrm{collision}}(x),
\end{equation}
where the semantic benefit is balanced against physical safety.

The navigation heatmap $H_{\mathrm{nav}}$ is defined as a distance field over traversable regions. Non-traversable pixels are set to zero, while values on free-space pixels decrease monotonically with distance to the navigation target and are normalized to $[0,1]$. This design ensures that the peak response lies at a reachable standing point adjacent to the object, rather than at the geometric center of the object mask. The facing heatmap $H_{\mathrm{fac}}$ is defined as a 2D Gaussian centered at the geometric center of the facing-target mask, with a peak value of one that decays smoothly outward. This converts discrete orientation selection into continuous spatial probability estimation, providing complementary supervision for downstream control (e.g., aligning with a monitor or a user).

\subsection{Heatmap Prediction Model}
\begin{figure}[t]
  \centering
  \includegraphics[width=0.95\linewidth]{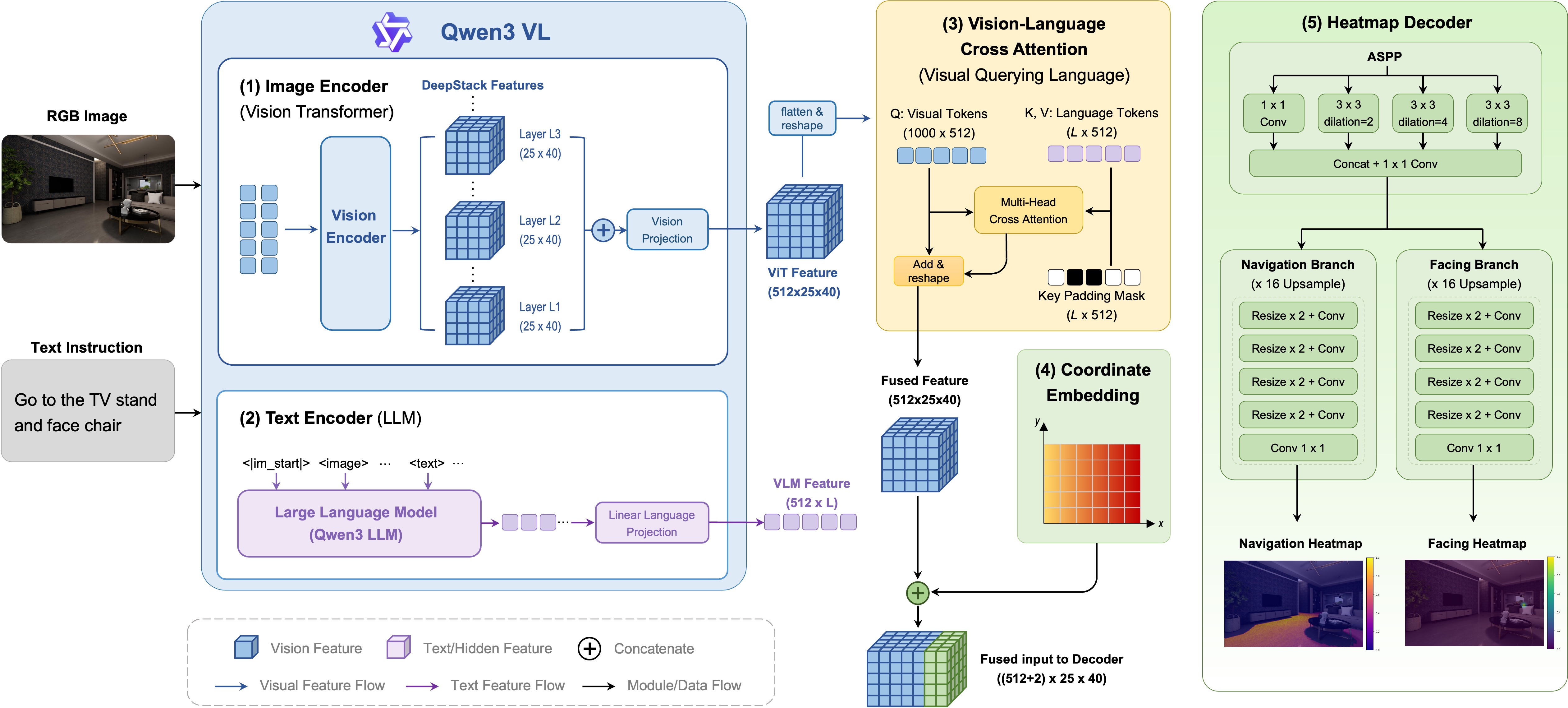}
  \caption{Overview of the proposed heatmap prediction model.}
  \label{fig:model_structure}
\end{figure}
\textbf{Overall architecture.} As illustrated in Figure~\ref{fig:model_structure}, we adopt a dual-stream design consisting of a VLM backbone, a vision-language cross-attention module, and a lightweight dense decoder that predicts both $H_{\mathrm{nav}}$ and $H_{\mathrm{fac}}$ via parallel convolutional heads:
\begin{equation}
(H_{\mathrm{nav}}, H_{\mathrm{fac}})=\mathrm{Decoder}(\mathrm{CrossAttn}(F_{\mathrm{vis}}, F_{\mathrm{vlm}}))
\end{equation}
Here $F_{\mathrm{vis}}$ denotes spatial visual features, and $F_{\mathrm{vlm}}$ denotes VLM context features.

\textbf{Backbone and feature extraction.} We use Qwen3-VL \cite{Qwen3-VL} as the backbone and apply LoRA \cite{hu2022lora} fine-tuning ($r=16$, $\alpha=32$) on all attention $q/v$ projections to preserve open-vocabulary capability with reduced memory overhead. For visual encoding, we concatenate three deepstack feature levels from the ViT encoder and project them into the decoder hidden space to obtain $F_{\mathrm{vis}}$. For VLM-context encoding, we feed the full multimodal dialogue sequence into the VLM decoder and project the resulting contextual features with a linear layer into the same hidden space to obtain $F_{\mathrm{vlm}}$.

\textbf{Vision-language cross-attention.} To preserve spatial grounding, we fuse modalities with visual tokens as queries and VLM tokens as keys/values, followed by an FFN residual block. A masking mechanism $\mathbf{M}$ suppresses padded tokens, image tokens, and assistant-response tokens to avoid information leakage, forcing the cross-attention to rely strictly on the user instruction semantics.

\textbf{Coordinate-aware decoder.} Inspired by CenterNet \cite{zhou2019objects}, we concatenate normalized coordinate channels ($x,y\in[-1,1]$) with the fused feature to form $\tilde{F}\in\mathbb{R}^{B\times(d+2)\times h_f\times w_f}$. The decoder utilizes a $1\times1$ bottleneck and ASPP with dilation rates $\{1,2,4,8\}$ for multi-scale aggregation. It then splits into two parallel convolutional branches (navigation and facing), each with four Resize-Conv upsampling blocks (bilinear $\times2$ + $3\times3$ convolution), outputting one logit map per branch. Resize-Conv is explicitly chosen over transposed convolution to eliminate checkerboard artifacts near heatmap peaks.

\subsection{Automated Synthetic Data Generation}
\begin{figure}[t]
  \centering
  \includegraphics[width=0.95\linewidth]{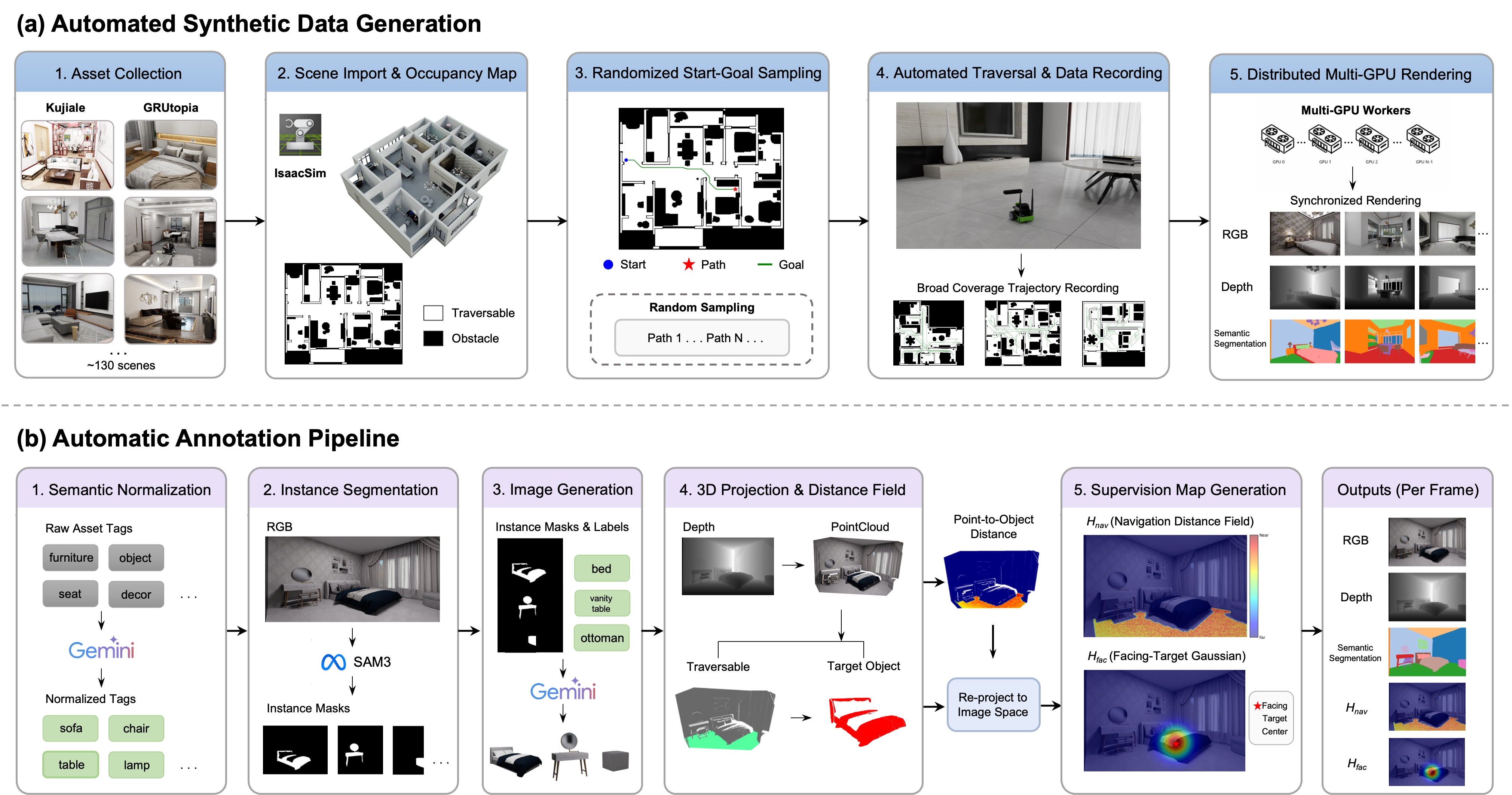}
  \caption{Overview of the automated synthetic data generation pipeline.}
  \label{fig:data_pipeline}
\end{figure}
To train the dense prediction model, we establish a fully automated synthetic data pipeline (Figure~\ref{fig:data_pipeline}). We collect high-quality assets from open datasets such as Kujiale \cite{InteriorAgent2025} and GRUtopia \cite{grutopia} (approx. 130 scenes) and import them into Isaac Sim, where an occupancy-map generator produces 2D traversability maps. The robot systematically traverses each scene, utilizing distributed multi-GPU rendering to obtain synchronized RGB, depth, and semantic-segmentation frames.

To eliminate manual annotation costs, we employ a foundation-model-assisted labeling pipeline. Gemini normalizes inconsistent asset tags (e.g., refining ``furniture'' to ``sofa''), and SAM3 extracts instance-level masks. Utilizing the depth ground truth, we project traversable regions and target objects into 3D point clouds, compute point-to-object distances, normalize them into a distance field, and re-project them to the 2D image space to generate the ground-truth $H_{\mathrm{nav}}$. A 2D Gaussian is placed at the facing-target mask center to obtain the ground-truth $H_{\mathrm{fac}}$. Masked object crops are further re-rendered by Gemini under diverse viewpoints to generate reference image patches, yielding large-scale, paired multimodal instruction supervision.

Beyond the virtual data rendered by our own simulation pipeline, we further collect ScanNet\cite{dai2017scannet}, SunRGBD\cite{song2015sun}, HyperSim\cite{roberts2021hypersim}, and Matterport3D\cite{Matterport3D}, and annotate them for our semantic-navigation setting. Since the benchmark composition is important for interpreting the generalization results, we defer the exact sample counts and train/test split across these data sources to Appendix~C.

\subsection{Training Objective} 
The overall loss function is defined as a weighted sum of the navigation and facing objectives:
\begin{equation}
\mathcal{L}=w_{\mathrm{nav}}\mathcal{L}_{\mathrm{nav}}+w_{\mathrm{fac}}\mathcal{L}_{\mathrm{fac}}
\end{equation}
Supervision is applied in the decoder-resolution space $(h_f\times w_f)$ by downsampling the ground-truth heatmaps. Empirically, balancing the navigation and facing tasks under a fixed weighting scheme is challenging. We adopt a dynamic annealing strategy for $w_{\mathrm{nav}}$ and $w_{\mathrm{fac}}$ during training. Gradually reducing $w_{\mathrm{fac}}$ toward the end of training mitigates late-stage optimization conflicts, yielding a superior compromise between reachable waypoint localization and orientation prediction.

For the navigation branch, we apply Binary Cross-Entropy (BCE) on the logits:
\begin{equation}
\mathcal{L}_{\mathrm{nav}}=-\frac{1}{N}\sum_{xy}\left[H_{xy}\log \hat{p}_{xy}+(1-H_{xy})\log(1-\hat{p}_{xy})\right]
\end{equation}
Compared to Mean Squared Error (MSE), BCE imposes a stronger penalty on confident false positives in non-traversable background regions.

For the facing branch, we employ a Gaussian-reweighted focal loss inspired by CenterNet:
\begin{equation}
\mathcal{L}_{\mathrm{fac}} = -\frac{1}{N_{\mathrm{pos}}}\sum_{xy}
\begin{cases}
(1-\hat{p}_{xy})^{\alpha}\log(\hat{p}_{xy}), & g_{xy}\ge 0.5 \\
(1-g_{xy})^{\beta}\hat{p}_{xy}^{\alpha}\log(1-\hat{p}_{xy}), & g_{xy}<0.5
\end{cases}
\end{equation}
with $\alpha=2$ and $\beta=4$. This effectively suppresses easily classified background pixels while preserving soft-Gaussian gradients near the target center.

\subsection{System Integration: Trajectory Grounding via Semantic Heatmaps}
To convert the predicted heatmaps into executable actions, our system decouples dense visual-semantic prediction from embodiment-aware action realization, avoiding the brittleness of regressing a single hard coordinate.

First, the predicted $H_{\mathrm{nav}}$ and $H_{\mathrm{fac}}$ are projected from the 2D image space into the robot-centered 3D geometric representation using the corresponding depth stream and the camera's extrinsic $T$ matrix. To rigorously bound spatial uncertainty during this cross-modal projection, the statistical estimation of the $T$ matrix error (both maximum and minimum deviations) is strictly computed using absolute values rather than raw signed values. This ensures that the worst-case geometric drift in the physical space is accurately accounted for when mapping the affordance fields.

Once projected, the heatmaps function as local semantic potential fields. For a given robot $\mathcal{R}$, we deploy a sampling-based local planner—instantiated in our system via a Model Predictive Path Integral (MPPI) algorithm. The planner rolls out candidate trajectories $\tau$ under embodiment-specific kinematic constraints. Rather than tracking a discrete waypoint, the planner evaluates each feasible, collision-free trajectory by integrating the semantic reward along its path:
\begin{equation}
S(\tau)=\int_{0}^{T} \left( \alpha \mathcal{M}_{\mathrm{nav}}(p_t) + \beta \mathcal{D}_{\mathrm{fac}}(\theta_t) \right) dt,
\end{equation}
where $\mathcal{M}_{\mathrm{nav}}$ is the projected spatial reward from $H_{\mathrm{nav}}$, and $\mathcal{D}_{\mathrm{fac}}$ aligns the final pose $\theta$ with the orientation constraints from $H_{\mathrm{fac}}$. The highest-scoring trajectory is dispatched to the low-level controller. This formulation allows the identical VLM output to be natively reused across diverse embodiments—ranging from wheel-based platforms to legged robots—while the downstream planner independently handles physical footprint constraints and locomotion primitives.

\section{Experiments}

Unless otherwise specified, all models reported in our experiments are trained on our generated simulation dataset. We provide the detailed data split across the constituent data sources in Appendix~C.

\subsection{Waypoint Prediction Benchmark}
We evaluate waypoint prediction on the MP3D \cite{Matterport3D} \texttt{val\_test} split and compare our method with several recent multimodal baselines, including InternVLA-N1-System2 \cite{internvla-n1}, Molmo \cite{deitke2024molmo}, and Qwen3-VL-Instruct \cite{Qwen3-VL}. All models use approximately 8B parameters for a fair zero-shot comparison. Unless otherwise specified, \textit{Ours} denotes our full model with dynamic facing-loss annealing. Our benchmark contains two coupled sub-tasks: navigation waypoint prediction and facing prediction. Navigation evaluates whether the model predicts a reachable point close to the target standing position, while facing evaluates whether the model correctly predicts the orientation-aligned target region. We additionally report overall metrics that jointly summarize both aspects.

\textbf{Metrics.} We extract the predicted point as the maximum-value location in the heatmap if its confidence exceeds a predefined threshold; otherwise, the model is treated as making no prediction. For navigation, a sample is counted as positive when a valid ground-truth waypoint exists. A true positive (TP) means the model outputs a prediction that lies on traversable ground and within a distance threshold of the ground-truth point. A false positive (FP) means the model outputs a point when no valid ground-truth waypoint exists, or outputs a point that is not on traversable ground or is outside the threshold. A false negative (FN) means a valid ground-truth waypoint exists but the model fails to output a prediction. For facing prediction, TP means that the predicted point falls inside the target-object mask. Based on these definitions, we report precision, recall, and F1-score. We further report Affordance Rate (AR), which measures the fraction of positive samples whose predicted navigation points lie in physically reachable regions, and MSE, which measures the average distance between the predicted point and the ground-truth point conditioned on the prediction being reachable. In all experiments, we set the confidence threshold to 0.5 and the distance threshold to 1.0\,m. Unlike global VLN tasks that heavily rely on Success weighted by Path Length (SPL) to evaluate long-horizon routing efficiency, our in-FOV semantic anchoring task prioritizes absolute physical reachability and local execution success; therefore, SR, AR, and local distance errors are our primary indicators.

\begin{table*}[t]
\centering
\scriptsize
\setlength{\tabcolsep}{2.4pt}
\caption{Waypoint prediction benchmark on the MP3D \texttt{val\_test} split. ``--'' indicates that the corresponding model does not provide facing predictions. Our method achieves the best zero-shot overall performance and substantially improves navigation recall over general-purpose VLM baselines.}
\label{tab:benchmark}
\resizebox{\textwidth}{!}{%
\begin{tabular}{lcccccccccccc}
\toprule
Model & Params & Nav P$\uparrow$ & Nav R$\uparrow$ & Nav F1$\uparrow$ & AR$\uparrow$ & MSE$\downarrow$ & Facing P$\uparrow$ & Facing R$\uparrow$ & Facing F1$\uparrow$ & Overall P$\uparrow$ & Overall R$\uparrow$ & Overall F1$\uparrow$ \\
\midrule
InternVLA-N1-System2\cite{internvla-n1} & 8B & 0.1408 & 0.0240 & 0.0410 & 0.0622 & 1.6194 & -- & -- & -- & -- & -- & -- \\
Molmo\cite{deitke2024molmo} & 8B & 0.1822 & 0.2246 & 0.2012 & 0.2492 & 1.0774 & 0.5817 & 0.4150 & 0.4844 & 0.3820 & 0.3453 & 0.3627 \\
Qwen3-VL-Instruct\cite{Qwen3-VL} & 8B & 0.2801 & 0.2044 & 0.2363 & 0.2138 & \textbf{0.9012} & 0.6090 & 0.3842 & 0.4723 & \textbf{0.4685} & 0.3137 & 0.3758 \\
\textbf{Ours} & 8B & \textbf{0.2999} & \textbf{0.6740} & \textbf{0.4151} & \textbf{0.7265} & 1.5617 & \textbf{0.6840} & \textbf{0.4616} & \textbf{0.5512} & 0.4645 & \textbf{0.5224} & \textbf{0.4917} \\
\bottomrule
\end{tabular}%
}
\end{table*}

Table~\ref{tab:benchmark} shows that our zero-shot model clearly outperforms the compared 8B baselines on navigation and overall performance. In particular, the large gain in navigation recall indicates that the proposed heatmap formulation is much more effective than directly regressing isolated points when transferring across embodiments and environments. Our framework also achieves the best overall zero-shot F1-score, suggesting that dense waypoint prediction provides a stronger inductive bias for grounding multimodal instructions into physically executable local goals.

\subsection{Feature Fusion Study}
We conduct a feature fusion study to examine how visual tokens from the ViT encoder should be integrated with language-conditioned representations. Specifically, we compare five variants: directly using the backbone VLM features without extra fusion, direct addition of ViT and VLM features, FiLM-based linear modulation, cross-attention between directly encoded text features and ViT features, and cross-attention between ViT features and VLM features. All variants in this study are trained and evaluated on our generated simulation dataset under the same protocol.

\begin{table*}[t]
\centering
\scriptsize
\setlength{\tabcolsep}{2.4pt}
\caption{Feature fusion study. We compare five ways of combining visual and language information: using the backbone VLM features directly, direct addition, FiLM-based modulation, cross-attention between directly encoded text and ViT features, and cross-attention between ViT and VLM features.}
\label{tab:feature_fusion}
\resizebox{\textwidth}{!}{%
\begin{tabular}{lccccccccccc}
\toprule
Fusion Variant & Nav P$\uparrow$ & Nav R$\uparrow$ & Nav F1$\uparrow$ & AR$\uparrow$ & MSE$\downarrow$ & Facing P$\uparrow$ & Facing R$\uparrow$ & Facing F1$\uparrow$ & Overall P$\uparrow$ & Overall R$\uparrow$ & Overall F1$\uparrow$ \\
\midrule
VLM feature only & 0.2743 & 0.7664 & 0.4041 & 0.8634 & 1.6434 & 0.3154 & 0.0245 & 0.0455 & 0.2769 & 0.2436 & 0.2592 \\
ViT + VLM (Direct Add) & 0.2796 & \textbf{0.9987} & 0.4335 & \textbf{0.9666} & 1.6832 & 0.5939 & 0.4885 & 0.5361 & 0.3872 & 0.6413 & 0.4828 \\
ViT + VLM (FiLM) & 0.2851 & 0.9904 & 0.4428 & 0.9572 & 1.8239 & 0.6085 & 0.3425 & 0.4383 & 0.3796 & 0.5252 & 0.4407 \\
ViT + Text (Cross-Attn) & 0.2650 & 0.9983 & 0.4188 & 0.9632 & 1.7296 & 0.3535 & 0.0254 & 0.0475 & 0.2698 & 0.2677 & 0.2687 \\
ViT + VLM (Cross-Attn) & \textbf{0.3641} & 0.9717 & \textbf{0.5297} & 0.9567 & \textbf{1.5685} & \textbf{0.7287} & \textbf{0.6051} & \textbf{0.6612} & \textbf{0.5165} & \textbf{0.7159} & \textbf{0.6001} \\
\bottomrule
\end{tabular}%
}
\end{table*}

Table~\ref{tab:feature_fusion} leads to three key observations. First, directly using the backbone VLM features is not competitive overall: although it provides a reasonable initialization, it is worse than all ViT+VLM fusion variants on the main navigation and overall metrics, and it is especially weak on facing prediction. This result indicates that the multimodal representation inside Qwen3-VL is useful, but directly decoding from it is insufficient for precise heatmap grounding.

Second, explicitly re-introducing ViT spatial features is clearly beneficial. Both direct addition and FiLM improve the model over using VLM features alone, with especially large gains in AR and overall performance. In particular, direct addition attains the best Nav R and the best AR, showing that enhancing the prediction head with explicit visual structure helps the model place targets inside executable free space much more reliably.

Third, the best overall trade-off is achieved by cross-attention between ViT features and VLM features. This ViT+VLM cross-attention variant achieves the best Nav P, Nav F1, MSE, all facing metrics, and all overall metrics, indicating that it is the most effective fusion strategy in our study. By contrast, replacing VLM features with directly encoded text features in the cross-attention module causes a large drop, especially on facing prediction and overall F1. This gap shows that the vision-language fusion already formed inside the Qwen3-VL backbone is indeed valuable for the downstream task. At the same time, the fact that both direct addition and ViT+VLM cross-attention outperform the VLM-only baseline confirms that simply consuming fused backbone tokens is still inferior to using them to enhance explicit visual features.

\subsection{Simulation Robot Validation}
Beyond offline evaluation, we also validate the proposed framework in robot simulation with three embodiments: Jetbot, H1, and Aliengo. For each robot, we report results on both seen and unseen scenes, where \emph{seen} denotes the training split of our generated simulation dataset and \emph{unseen} denotes its test split. We evaluate two metrics. \textbf{Success Rate (SR)} measures the fraction of episodes in which the model predicts a valid navigation point and the robot successfully moves to within a threshold of the ground-truth point. \textbf{Navigation Error (NE)} measures the average distance between the robot's final position and the ground-truth position over the episodes where the model outputs a valid navigation point. In all simulation experiments, the success threshold is set to 1.0\,m.

\begin{table*}[t]
\centering
\scriptsize
\setlength{\tabcolsep}{2.4pt}
\caption{Simulation robot validation across different embodiments. SR denotes Success Rate and NE denotes Navigation Error. For each robot, we report performance on both seen and unseen scenes.}
\label{tab:sim_robot_validation}
\resizebox{\textwidth}{!}{%
\begin{tabular}{lcccccccccccc}
\toprule
& \multicolumn{4}{c}{Jetbot} & \multicolumn{4}{c}{H1} & \multicolumn{4}{c}{Aliengo} \\
\cmidrule(lr){2-5} \cmidrule(lr){6-9} \cmidrule(lr){10-13}
Method & \multicolumn{2}{c}{Seen} & \multicolumn{2}{c}{Unseen} & \multicolumn{2}{c}{Seen} & \multicolumn{2}{c}{Unseen} & \multicolumn{2}{c}{Seen} & \multicolumn{2}{c}{Unseen} \\
\cmidrule(lr){2-3} \cmidrule(lr){4-5} \cmidrule(lr){6-7} \cmidrule(lr){8-9} \cmidrule(lr){10-11} \cmidrule(lr){12-13}
 & SR$\uparrow$ & NE$\downarrow$ & SR$\uparrow$ & NE$\downarrow$ & SR$\uparrow$ & NE$\downarrow$ & SR$\uparrow$ & NE$\downarrow$ & SR$\uparrow$ & NE$\downarrow$ & SR$\uparrow$ & NE$\downarrow$ \\
\midrule
Qwen3-VL & 0.4146 & \textbf{0.1965} & 0.3686 & \textbf{0.6640} & 0.2175 & \textbf{0.6169} & 0.2200 & \textbf{0.9790} & 0.1585 & \textbf{1.2469} & 0.1548 & 1.5280 \\
\textbf{Ours} & \textbf{0.7012} & 0.8890 & \textbf{0.6517} & 0.9275 & \textbf{0.6004} & 1.7038 & \textbf{0.5693} & 1.0952 & \textbf{0.5230} & 1.5981 & \textbf{0.4790} & \textbf{1.3377} \\
\bottomrule
\end{tabular}%
}
\end{table*}

Table~\ref{tab:sim_robot_validation} shows that our model achieves substantially higher success rates than the fine-tuned Qwen3-VL baseline across all three robot embodiments and in both seen and unseen scenes. This consistent SR gain is the clearest evidence of cross-embodiment transfer: predicting a dense navigation heatmap enables the system to select targets that are more likely to lie in executable free space, and is therefore better aligned with the core requirement of embodied navigation. 

By contrast, the NE results should be interpreted with care due to a prominent \textbf{survivorship bias}. In our protocol, NE is computed only for episodes in which the model predicts a point that can actually be executed by the robot. As a result, the baseline Qwen3-VL excludes a massive number of failure cases where its predicted points fall into non-traversable space (e.g., inside obstacles). The remaining valid predictions are heavily biased toward a filtered subset that happens to lie closer to the ground-truth point. This survivorship bias artificially lowers its NE, even though its overall navigation reliability is severely compromised. Therefore, the combination of much higher SR and non-uniform NE strongly suggests that predicting a single sparse waypoint is significantly less robust for embodied execution than predicting a heatmap distribution over reachable regions.

\section{Conclusion}

We present a Dual-Heatmap framework for in-FOV semantic navigation that maps open-vocabulary multimodal intent to physically reachable robot actions. Instead of regressing a single brittle waypoint, the model predicts navigation and facing heatmaps that act as semantic potential fields for downstream planning.

Built on a fully automated data-generation pipeline, the method achieves state-of-the-art zero-shot performance among comparable 8B baselines. Our feature-fusion and simulation results further show that dense heatmap prediction improves affordance grounding, mitigates the survivorship bias of point regression, and transfers across different robot embodiments. We will release the dataset, generation pipeline, and simulation environments to support reproducible embodied-navigation research.

\bibliographystyle{abbrvnat}
\bibliography{ref}

\newpage
\appendix
\section*{Appendix}
\addcontentsline{toc}{section}{Appendix}

\subsection*{A. Qualitative Results}
To intuitively demonstrate the advantages of the proposed Dual-Heatmap representation over single-point regression, we visualize qualitative comparisons in complex indoor scenarios. 

\begin{figure}[ht]
  \centering
  \includegraphics[width=0.95\linewidth]{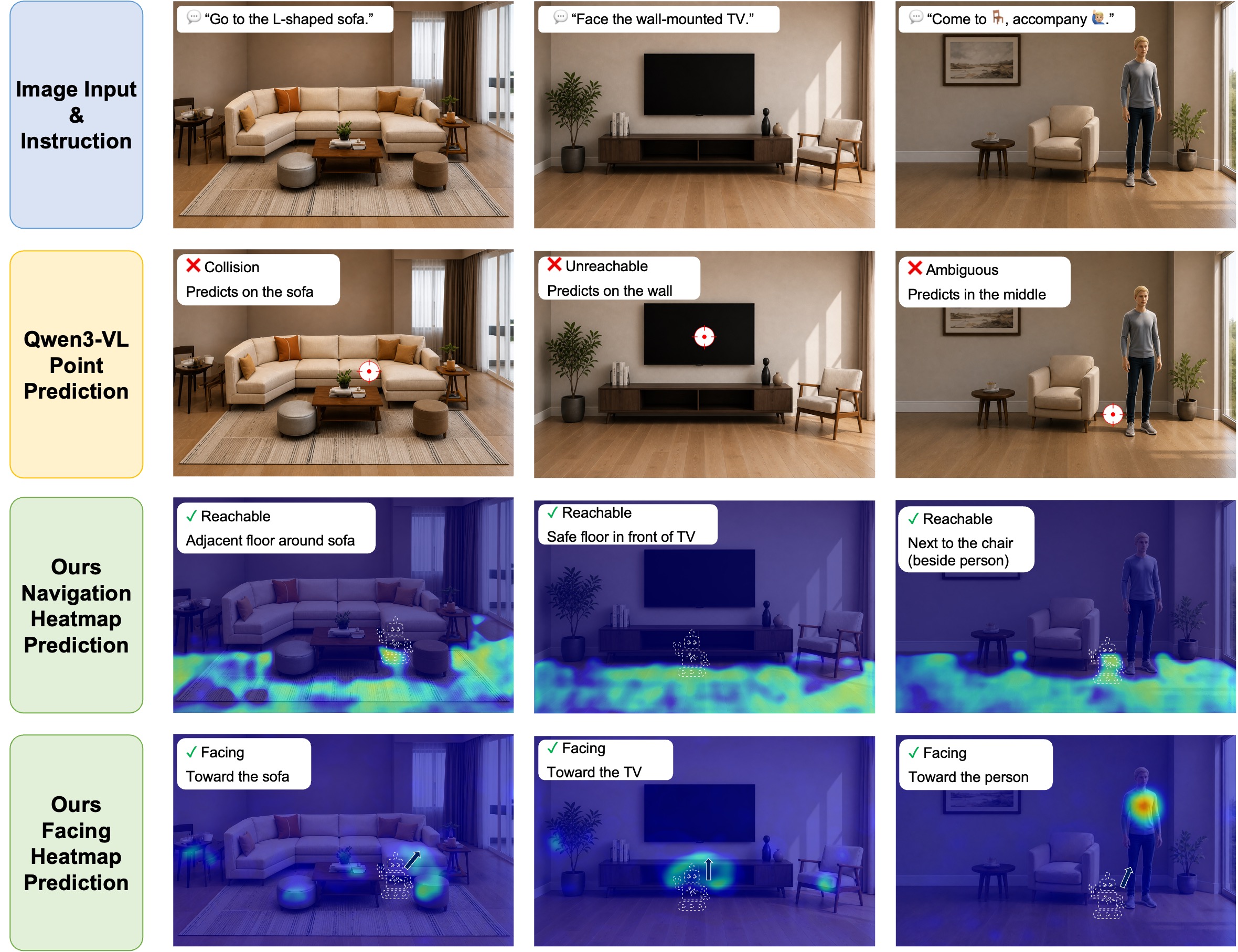}
  \caption{Qualitative comparison of spatial grounding. Given a multimodal instruction, the baseline point-regression model (red cross) often predicts a coordinate that collides with the furniture. In contrast, our model predicts a continuous affordance heatmap ($H_{\mathrm{nav}}$) covering the reachable adjacent floor, and correctly identifies the interaction orientation ($H_{\mathrm{fac}}$).}
  \label{fig:qualitative_heatmap}
\end{figure}

\begin{figure}[ht]
  \centering
  \includegraphics[width=0.75\linewidth]{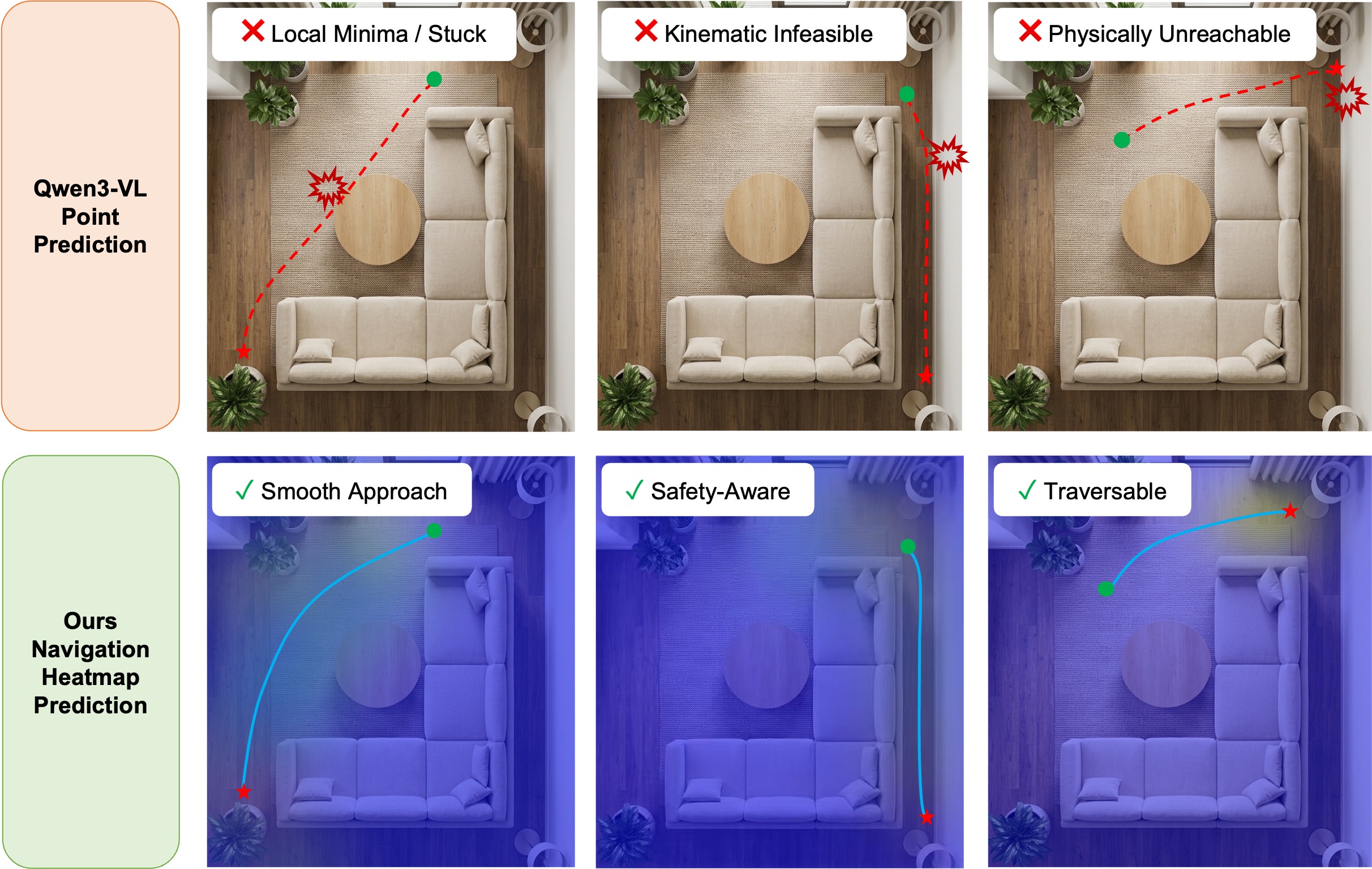}
  \caption{Execution trajectories in the simulation environment. While deterministic waypoints easily trap the robot in local minima near obstacles (dashed red line), our semantic potential field seamlessly guides the MPPI planner along a collision-free and physically smooth trajectory (solid blue line) to the optimal standpoint.}
  \label{fig:qualitative_trajectory}
\end{figure}

As shown in Figure~\ref{fig:qualitative_heatmap}, when targeting an object with irregular boundaries, point-regression models often output a coordinate that geometrically makes sense but is physically non-traversable. Our $H_{\mathrm{nav}}$ explicitly highlights the topological free space, preserving multiple valid interaction standpoints. Furthermore, Figure~\ref{fig:qualitative_trajectory} illustrates the system integration. By treating the heatmap as a semantic potential field rather than a rigid goal, the downstream local planner dynamically avoids unforeseen obstacles while naturally flowing toward the highest affordance region, ensuring both semantic compliance and physical safety.

\subsection*{B. Reproducibility and Open Benchmark}
A persistent challenge in embodied AI research is the lack of standardized, reproducible benchmarks, as real-world physical evaluations are notoriously difficult to replicate across different laboratories due to hardware and environmental discrepancies. Rather than relying on isolated real-world robot demonstrations that cannot be easily verified, a core contribution of our work is the establishment of a unified, open-source simulation benchmark. 

Upon publication, we will publicly release our fully automated synthetic data generation pipeline, the large-scale multimodal instruction dataset, and the integrated ROS2-based simulation testing environments spanning multiple robot embodiments (Jetbot, H1, Aliengo). We believe that providing a transparent, highly reproducible platform will better facilitate fair comparisons and accelerate future research in fine-grained visual-language navigation and spatial affordance grounding.

\subsection*{C. Dataset Details}
To make the data foundation of our model explicit, we summarize here the composition of the dataset described in Section~3.3. In addition to the virtual data rendered by our own pipeline, we collect ScanNet\cite{dai2017scannet}, SunRGBD\cite{song2015sun}, HyperSim\cite{roberts2021hypersim}, and Matterport3D\cite{Matterport3D} and annotate them for our task setting. In the main paper, we reference this appendix whenever discussing the training data distribution.

Table~\ref{tab:dataset_stats} summarizes the sample distribution of the training and test splits across all constituent datasets. In total, the corpus contains 266{,}766 training samples and 79{,}140 test samples, for 345{,}906 samples overall.

\begin{table}[h]
\centering
\scriptsize
\caption{Dataset statistics for the multimodal semantic-navigation corpus used to train and evaluate our model.}
\label{tab:dataset_stats}
\begin{tabular}{llr}
\toprule
Dataset & Split & Samples \\
\midrule
ScanNet\cite{dai2017scannet} & train & 40733 \\
 & test & 3266 \\
SunRGBD\cite{song2015sun} & train & 10887 \\
 & test & 10971 \\
HyperSim\cite{roberts2021hypersim} & train & 124424 \\
 & test & 32587 \\
Matterport3D\cite{Matterport3D} & train & 59878 \\
 & test & 22343 \\
Ours & train & 30844 \\
 & test & 9973 \\
\midrule
All datasets & train & 266766 \\
 & test & 79140 \\
 & total & 345906 \\
\bottomrule
\end{tabular}
\end{table}


\end{document}